\documentclass{article}
\usepackage{spconf,amsmath,graphicx}
\usepackage{amssymb}
\usepackage{booktabs}
\usepackage{multirow}
 \usepackage{cite}
 \usepackage{hyperref}

\usepackage{todonotes}

\title{privacy attacks  for automatic speech recognition acoustic models in  a Federated learning framework }

\name{Natalia Tomashenko$^1$, Salima Mdhaffar$^1$, Marc Tommasi$^2$,  Yannick Estève$^1$, Jean-François Bonastre$^1$
\thanks{This work was funded  by VoicePersonae and DEEP-PRIVACY  (ANR-18-CE23-0018) projects. This work was performed using HPC resources from GENCI-IDRIS (Grant 2021-AD011013331).}
}
\address{$^1$ LIA, Avignon Université, France \\
$^2$ Université de Lille, CNRS, Inria, Centrale Lille, UMR 9189 - CRIStAL, Lille, France}
\begin{document}
\ninept
\maketitle
\begin{abstract}

This paper investigates methods to effectively retrieve  speaker information  from the personalized speaker adapted neural network  acoustic models (AMs) in automatic speech recognition (ASR).
This problem is especially important in the context of  federated learning of ASR acoustic models  where 
a  global model is learnt on the server based on the updates received from  multiple clients.
We propose an approach to analyze information in  neural network AMs based on a
neural network footprint on the so-called \textit{Indicator} dataset.  
Using  this method, we develop two attack models that aim to infer speaker identity from the  updated personalized models without  access to the actual users' speech data. 
Experiments on the TED-LIUM 3 corpus demonstrate that  the proposed approaches are very effective and can provide equal error rate (EER) of $1$--$2\%$.

\end{abstract}
\begin{keywords}
Privacy, federated learning, acoustic models, attack models,  speech recognition,  speaker verification
\end{keywords}
\section{Introduction}
\label{sec:intro}

Federated learning  (FL) for automatic speech  recognition (ASR) has recently become an active area of research~\cite{cui2021federated,dimitriadis2020federated,mdhaffar2021study,guliani2021training,dimitriadis2020dynamic,yu2021federated}.
To preserve the privacy of the users' data in the FL framework, the model is updated in a distributed fashion instead of communicating the data directly from clients to a server.

Privacy is one of the major challenges in FL~\cite{li2020federated,mothukuri2021survey}.
Sharing model updates, i.e. gradient information, instead of raw user data aims to protect user personal data that are processed locally on devices. However, these updates may still reveal some sensitive information
to a server or to a third party~\cite{geiping2020inverting,carlini2019secret}. 
According to recent research, FL has various privacy risks and may be vulnerable to different types of attacks, i.e. membership inference attacks~\cite{truex2019demystifying} or generative adversarial network (GAN) inference attacks~\cite{wang2019beyond}.
Techniques to enhance the privacy in a FL framework are mainly based on two categories~\cite{mothukuri2021survey}: secure multiparty computation~\cite{bonawitz2016practical} and differential privacy~\cite{dwork2006differential}.
Encryption methods~\cite{pathak2013privacy,smaragdis2007framework} 
such as fully homomorphic encryption~\cite{smaragdis2007framework} and secure multiparty computation perform computation  in the
encrypted domain.  These methods   increase  computational complexity.
In a FL framework, this increase is not so significant compared to standard centralized training, since  only the transmitted parameters   need to be encrypted instead of large amounts of  data, however with an increased number of participants, computational complexity becomes a critical issue.
Differential privacy methods  preserve privacy by adding noise to users' parameters~\cite{dwork2006differential,xie2018differentially}, however such solutions
may degrade learning performance due to the uncertainty they introduce into the parameters. 
Alternative methods to  privacy protection for speech include  deletion methods~\cite{cohen2019voice} that are meant for ambient sound analysis, and anonymization~\cite{tomashenko2020introducing,Tomashenko2021CSl} that aims to suppress personally identifiable information in the speech signal keeping unchanged all other attributes.
These privacy preservation methods can be combined and integrated  in a hybrid fashion into a FL framework.

 Despite the recent interest in FL for ASR and other speech-related tasks such as
 keyword spotting~\cite{leroy2019federated,hard2020training}, emotion recognition~\cite{latif2020federated}, and speaker verification~\cite{granqvist2020improving}, there is a lack of research on vulnerability of ASR acoustic models (AMs)
 to privacy attacks 
 in a FL framework. 
 In this work, we make a step towards this direction by analyzing  speaker information that can be retrieved from the 
 personalized  AM locally updated on the user's data. 
 To achieve this goal, we developed two privacy attack models that operate directly on the updated model parameters without access to the actual user's data.
 Parameters  of neural network (NN) personalized  AMs  contain a wealth  of  information about the  speakers~\cite{salima2022retrieving}. 
 In this paper, we  propose novel methods to efficiently and easily retrieve  speaker information from the adapted AMs.  The main idea of the proposed methods is to use  an external \textit{Indicator} dataset to analyze the footprint of AMs on this data.
 Another important contribution of this work is understanding how the speaker information is distributed in the adapted NN AMs.

 This paper is structured as follows.  Section~\ref{sec:fl} briefly introduces a considered FL framework for  AM training. Section~\ref{sec:attacks} describes the privacy preservation scenario and proposes two attack models.
 Experimental evaluation is presented in Section~\ref{sec:experiments}. We conclude in Section~\ref{sec:concl}.

\section{Federated learning for ASR acoustic models}
\label{sec:fl}

\begin{figure}[htb]
\centering
  \centerline{\includegraphics[width=8.52cm]{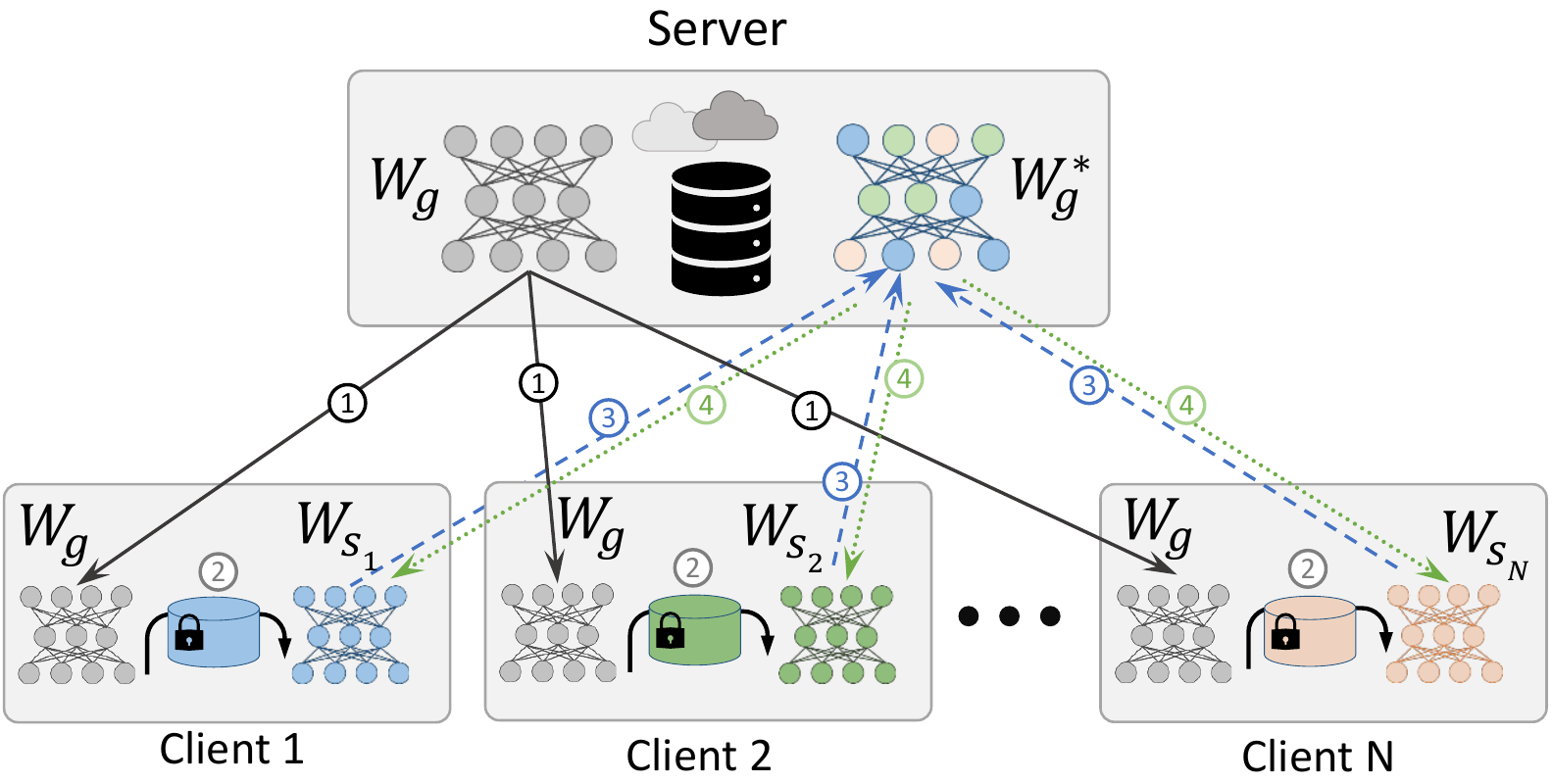}}
 \caption{Federated learning in a distributed network of clients:
 1)~Download of  the global model $W_g$ by clients.
 2) 
 Speaker adaptation of  $W_g$ on the local devices using user private  data.
 3)  Collection and aggregation of multiple personalized models $W_{s_1}$,...,$W_{s_N}$  on the sever.  4) Sharing the resulted model $W_g^*$ with the clients.}
\label{fig:fl}
\end{figure}

We consider a classical FL scenario where 
a global NN AM is trained on a server from the data stored locally on multiple remote devices~\cite{li2020federated}. The training of the global model is performed under the constraint that the training speech data are stored and processed locally on the user devices (clients), while only model updates are transmitted to the server from each client. The global model is learnt on the server based on the updates received from  multiple clients.
The FL in a distributed network of clients is illustrated in Figure~\ref{fig:fl}.
First, an  initial global speech recognition AM $W_g$
is distributed  to the group of devices of $N$ users (speakers).
Then, the initial  global model is run  on every user $s_i$ ($i\in1..N$) device and updated locally on the private user data. The updated models $W_{s_i}$ are then transmitted to the server where they  are aggregated to obtain a new global model $W_g^*$. 
Typically, the personalized updated models are aggregated using 
federated averaging and its variations~\cite{mcmahan2017communication,bonawitz2016practical}. 
Then, the updated global model $W_g^*$ is shared with the clients. 
The process restarts and loops until convergence or after a fixed number of rounds. 
The utility and training efficiency of the FL  AMs have been successfully studied in recent works~\cite{cui2021federated,dimitriadis2020federated,mdhaffar2021study,guliani2021training,dimitriadis2020dynamic,yu2021federated}, and these topics are beyond the scope of the current paper. Alternatively, we focus on the privacy aspect of this framework.


\section{Attack models}
\label{sec:attacks}

In this section, we 
describe the privacy preservation scenario
and present two attack models.

\subsection{Privacy preservation scenario}
\label{sec:pps}
Privacy preservation is formulated as a game between \textit{users} who share some data and \textit{attackers} who access this data or data derived from it and aim to infer information about the users~\cite{tomashenko2020introducing}.
To preserve the user data, in FL, there is no speech data exchange between a server and clients, only model updates are transmitted between the clients and server (or between some clients).  
Attackers aim to attack users using information owned by the server. 
They can get access to some updated personalized models.

In this work, we assume that an attacker 
 has access to the following data:
\begin{itemize}
    \item An initial global NN AM $W_g$;
    \item A personalized model $W_s$ of the  target speaker $s$ who is enrolled in the FL system. The corresponding personalized model was obtained from the global model $W_g$   by fine-tuning $W_g$ using  speaker data. 
    We consider this model as  \textit{enrollment} data for an attacker.
    \item Other personalized models of  non-target and target speakers: $W_{s_1}$,...,$W_{s_N}$. We will refer to these models as \textit{test trial} data.
\end{itemize}

The attacker's objective is to conduct an automatic speaker verification (ASV) task
 by using the enrollment data model in the form of  $W_s$ and test trial data in the form of models  $W_{s_1}$,...,$W_{s_N}$.

\subsection{Attack models}
\label{ssec:attacks}

The motivation of the proposed approaches is based on the hypothesis that we can capture  information about the identity of speaker $s$ from the corresponding speaker-adapted model $W_s$ and the global model $W_g$  
by comparing the outputs of these two neural AMs taken from  hidden layers $h$ on some speech data.
We will refer to this speech data as \textit{Indicator} data. Note, that the  \textit{Indicator}  data is not related to any test or AM training data and can be chosen arbitrarily from any speakers.  

\subsubsection{Attack model A1}
\label{sssec:a1}

\begin{figure}[htb]
\centering
  \centerline{\includegraphics[width=8.4cm]{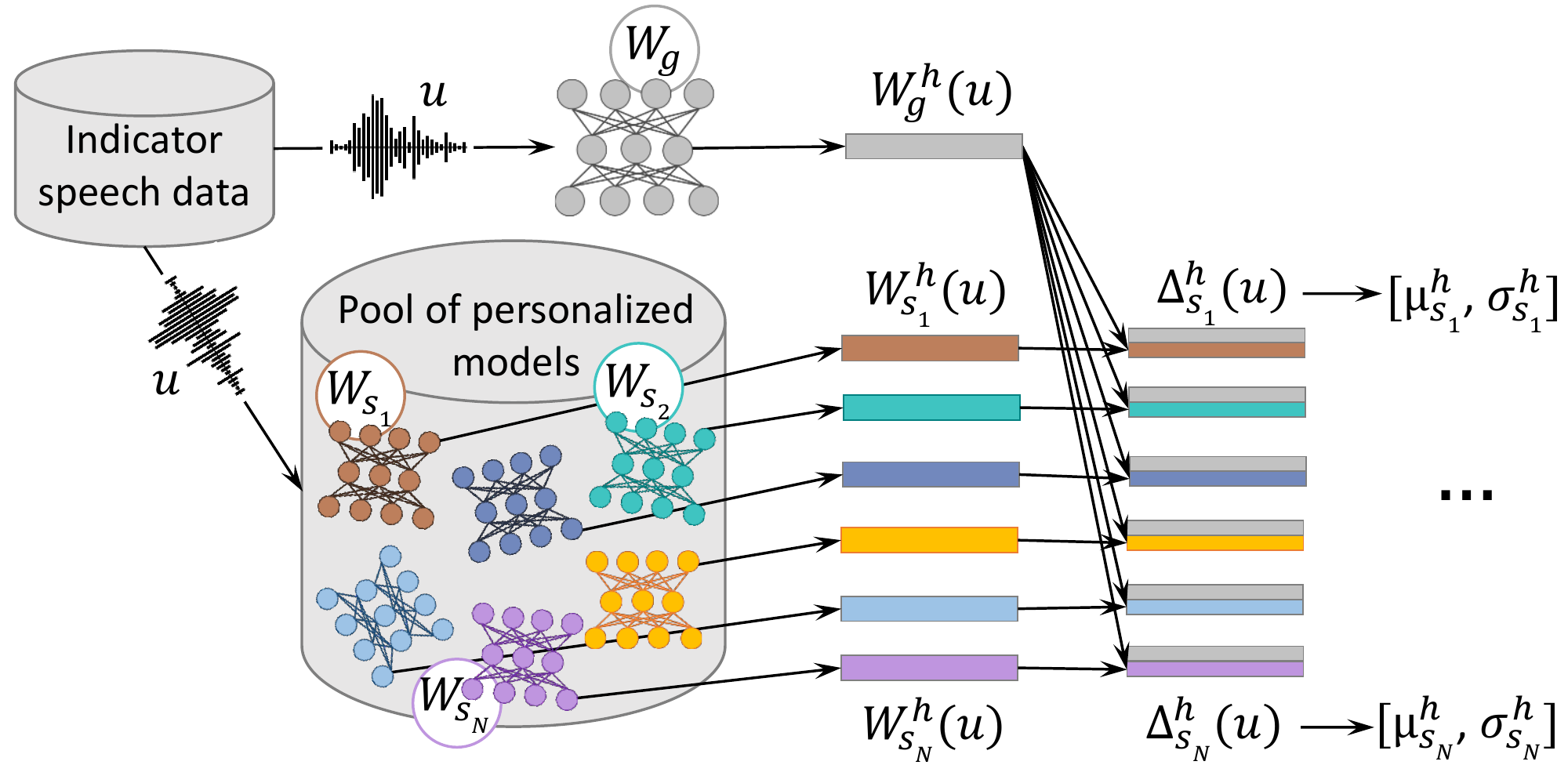}}
 \caption{Statistic computation for the attack model \textbf{A1}.}
\label{fig:a1}
\end{figure}

The ASV task with the proposed attack model is performed in several steps as illustrated in Figure~\ref{fig:a1}.
Let denote a set of  utterances in the  \textit{Indicator} dataset $\mathbb{I}$ as $\boldsymbol{u}_1,\ldots,\boldsymbol{u}_J\in \mathbb{I}$; 
a sequence of  vectors in utterance $\boldsymbol{u}_j$ as $\{u_j^{(1)}$, \ldots$,u_j^{(T_j)}\}$;
 a set of  personalized 
models as $W_{s_1},\ldots,W_{s_N}\in\mathbb{W}$; 
and  an identifier of a hidden layer in the global or personalized AM as $h$. 
\begin{enumerate}
    \item $\forall$ $W_{s_i}\in\mathbb{W}$, $\forall$ $\boldsymbol{u}_j\in\mathbb{I}$ we compute  activation values from the layer $h$ for model pairs: 
    $W_{s_i}^h(\boldsymbol{u}_j)=\{w_{s_i,j}^{h,t}\}_{t=1}^{T_j}$   and 
     $W_{g}^h(\boldsymbol{u}_j)=\{w_{g,j}^{h,t}\}_{t=1}^{T_j}$,  
    and per-frame differences between corresponding outputs:
       \begin{equation}
       \small
        \label{eq:delta}
    \boldsymbol{\Delta}_{s_i}^h(\boldsymbol{u}_j)=\{\Delta_{s_i,j}^{h,t}\}_{t=1}^{T_j},
     \end{equation}
      where $\Delta_{s_i,j}^{h,t}=w_{s_i,j}^{h,t}-w_{g,j}^{h,t}$, $t\in{1..T_j}$.
    \item For each personalized model, we compute mean and standard deviation vectors for $\Delta_{s_i,j}^{h,t}$ over all speech frames    in the \textit{Indicator} dataset $\mathbb{I}$:
\begin{equation}
\small
     \boldsymbol{\mu}_{s_i}^{h}=\frac{\sum_{j=1}^{I}{\sum_{t=1}^{T_j}\Delta_{s_i,j}^{h,t}}}{\sum_{j=1}^{I}{T_j}},
\end{equation}
 %
 \begin{equation}
 \small
      \boldsymbol{\sigma}_{s_i}^{h}=\left(\frac{\sum_{j=1}^{I}{   \sum_{t=1}^{T_j}(\Delta_{s_i,j}^{h,t}-\boldsymbol{\mu}_{s_i}^h)^2}}{\sum_{j=1}^{I}{T_j}}\right)^{\frac{1}{2}}
\end{equation}
\item For a pair of personalized models $W_{s_i}$ and $W_{s_k}$, we  compute a similarity score $\rho$ at hidden level $h$ on the \textit{Indicator} dataset based on the $\boldsymbol{L}_2$-normalised Euclidean distance between the corresponding vector  pairs for means and standard deviations:
%
\begin{equation}
\label{eq:rho}
    \rho(W_{s_i}^h, W_{s_k}^h)=\alpha_{\mu}\frac{\lVert \boldsymbol{\mu}_{s_i}^{h} - \boldsymbol{\mu}_{s_k}^{h} \rVert_2}{\lVert \boldsymbol{\mu}_{s_i}^{h}\rVert_2\lVert \boldsymbol{\mu}_{s_k}^{h}\rVert_2} + \alpha_{\sigma}\frac{\lVert \boldsymbol{\sigma}_{s_i}^{h} - \boldsymbol{\sigma}_{s_k}^{h} \rVert_2}{{
    \lVert \boldsymbol{\sigma}_{s_i}^{h}\rVert_2\lVert \boldsymbol{\sigma}_{s_k}^{h}\rVert_2}},
\end{equation}
where $\alpha_{\mu}$, $\alpha_{\sigma}$  are  fixed parameters 
in all experiments.
\item Given similarity scores for all matrix pairs, we can complete a speaker verification task based on these scores.
\end{enumerate}


\subsubsection{Attack model A2}
\label{sssec:a2}

\begin{figure}[htb]
\centering
  \centerline{\includegraphics[width=7.2cm]{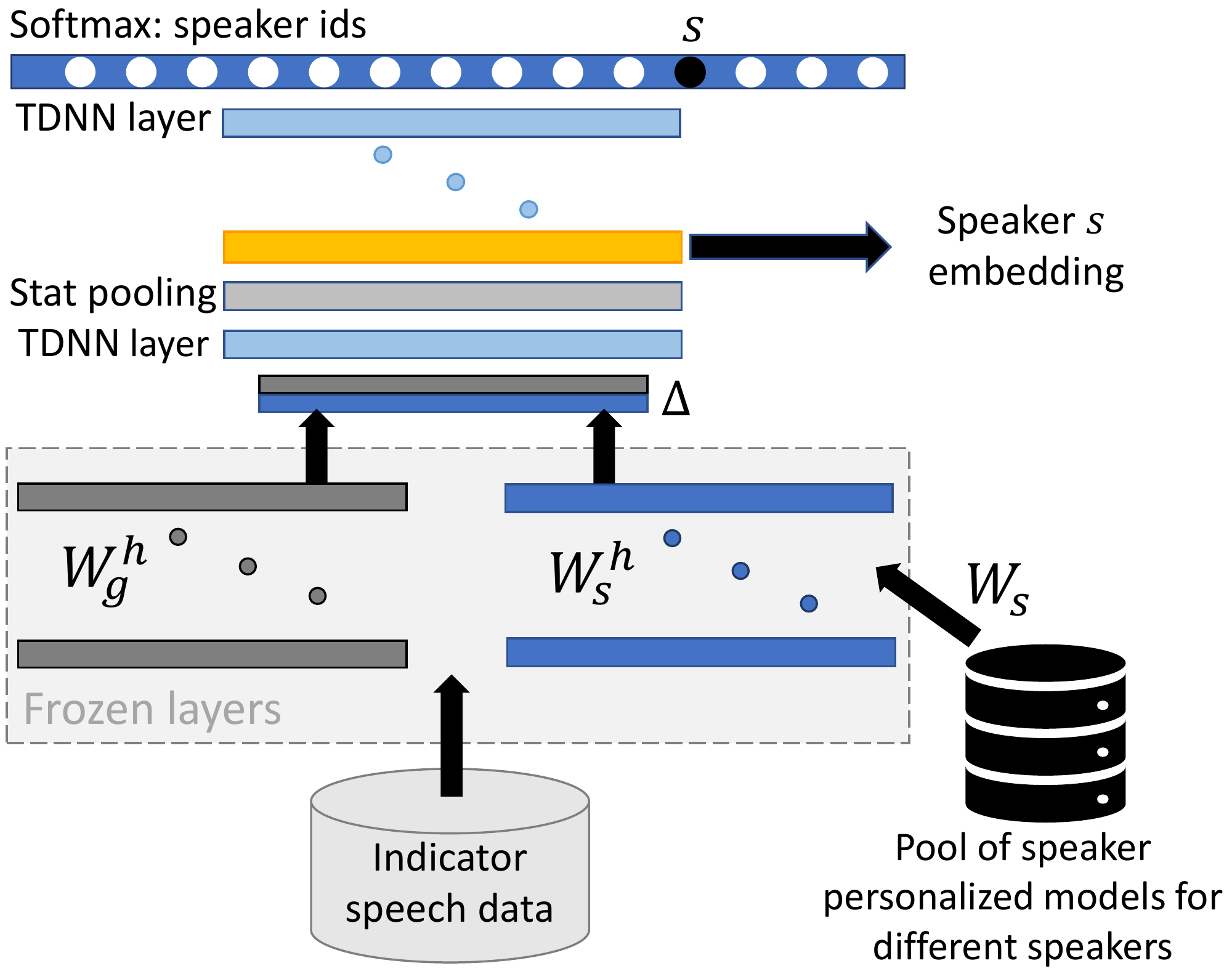}}
 \caption{Training a speaker embedding extractor for the attack model~\textbf{A2}.}
\label{fig:a2}
\end{figure}

For the second attack model, we train a NN model  as shown in Figure~\ref{fig:a2}.
This NN model uses  personalized and global models and the speech \textit{Indicator} dataset for training.
It is trained to predict a speaker identity provided the corresponding personalized model.
When the model is trained, we use it in  evaluation time to extract speaker embeddings similarly to x-vectors
and  apply probabilistic linear discriminant analysis (PLDA)~\cite{snyder2018x,ioffe2006probabilistic}.

As shown in Figure~\ref{fig:a2}, the model consists of two parts (frozen and trained).  The outputs of the frozen part are $\boldsymbol{\Delta}_{s_i}^h$ sequences of vectors computed per utterance of the \textit{Indicator} data as defined in Formula~(\ref{eq:delta}). 
For every  personalized model $W_{s_i}$, we  compute 
$\boldsymbol{\Delta}_{s_i}^h$ for all the utterances $\boldsymbol{u}$ of  the \textit{Indicator} corpus; then $\boldsymbol{\Delta}_{s_i}^h(\boldsymbol{u})$  is used as input to the second (trained) part of the NN which comprises several time delay neural network (TDNN) layers~\cite{peddinti2015time}  and one statistical pooling layer.

\vspace{-20pt}

\section{Experiments}
\label{sec:experiments}
\vspace{-1pt}

\subsection{Data}
\label{ssec:data}

The experiments were conducted on the   speaker adaptation partition of the TED-LIUM 3 corpus~\cite{hernandez2018ted}.  This
publicly available data set contains  TED talks that amount
to  452 hours  speech
data in English from about 2K speakers, 16kHz. 
Similarly to~\cite{mdhaffar2021study}, we selected from the TED-LIUM 3 training dataset three datasets: \textit{Train-G, Part-1, Part-2} with disjoint speaker subsets as shown in Table~\ref{tab:data}. The \textit{Indicator} dataset was used to train  an attack model. 
It is comprised of  320 utterances selected from all 32 speakers of test and development datasets of the TED-LIUM 3 corpus. 
The speakers in the \textit{Indicator} dataset are disjoint from speakers in \textit{Train-G, Part-1}, and \textit{Part-2}.
For each speaker in the \textit{Indicator} dataset we  select 10 utterances. The size of the \textit{Indicator} dataset is 32 minutes.  
The \textit{Train-G} dataset was used to train an initial global AM $W_g$.
\textit{Part-1} and \textit{Part-2} were used to obtain two sets of personalized models.\footnote{Data partitions and scripts will be available online: \url{https://github.com/Natalia-T/privacy-attacks-asr}}

 \begin{table}[ht]
  \centering
  \scalebox{0.9}{
  \renewcommand{\tabcolsep}{0.09cm}
  \begin{tabular}{l|l|l|l|l}
   \toprule
    & \textbf{Train-G}  &  \textbf{Part-1} &  \textbf{Part-2} & \textbf{Indicator}\\ \midrule
    Duration, hours &    200  &  86  &     73 & 0.5\\ 
    Number of speakers  &     880  & 736  & 634 & 32\\
    Number of personalized models & --- & 1300 & 1079 & --- \\
    \bottomrule
  \end{tabular}}
  \caption{Data sets statistics}
  \label{tab:data}
\end{table}
 \vspace{-8pt}

\subsection{ASR acoustic models}
\label{ssec:acm}

The ASR AMs have a 
TDNN model architecture~\cite{peddinti2015time}  and were trained using the Kaldi speech recognition toolkit~\cite{povey2011kaldi}.
 40-dimen\-sional Mel-frequency cepstral coefficients (MFCCs) without cepstral truncation  appended with 100-dimensional i-vectors were used as the input into the NNs.
 Each model has thirteen 512-dimensional hidden layers 
 followed by a softmax layer where 3664 triphone states were used as targets\footnote{Following the notation from \cite{peddinti2015time}, the model configuration can
be described as follows: \{-1,0,1\} $\times$ 6 layers; \{-3,0,3\} $\times$ 7 layers.}. 
The initial global model $W_g$ was trained using the lattice-free maximum mutual information (LF-MMI) criterion  with a 3-fold reduced frame rate as in \cite{povey2016purely}. 
The two types of speech data augmentation strategies were applied for the  training and adaptation data: speed perturbation (with factors 0.9, 1.0, 1.1)
and volume perturbation, as in~\cite{peddinti2015time}. Each model has about 13.8M parameters. The initial global model $W_g$ was trained on the \textit{Train-G}.
Personalized models $W_{s_i}$ were obtained by fine-tuning all the  parameters of $W_g$ on the speakers' data from \textit{Part-1} and \textit{Part-2} as described in~\cite{mdhaffar2021study}. For all personalized speaker models, we use approximately the same amount of speech data to perform fine-tuning (speaker adaptation) -- about 4 minutes per model.  For most of the speakers (564 in \textit{Part-1}, 463 in \textit{Part-2}) we obtained two  different  personalized models (per speaker) on disjoint adaptation subsets, for the rest speakers we have adaptation data only  for one model.

\subsection{Attack models}
\label{ssec:attacks-exp}
We investigate two approaches for attack models: \textbf{A1} -- a simple approach  based on the
comparative statistical analysis 
of the NN outputs and the associated   similarity score between personalized models,
and \textbf{A2} -- a NN based approach.
For  the test target trials, we use  comparisons between different personalized models  of the same speakers (564 in \textit{Part-2} and 1027 in the  \textit{Part-1+Part-2}), and for the non-target trials we randomly selected 10K pairs of models from different speakers in a corresponding dataset.

\vspace{-4pt}
\subsubsection{Attack model A1}
\label{sssec:a1e}

The first attack model was applied as described in Section~\ref{sssec:a1}. The parameters $\alpha_{\mu}$, $\alpha_{\sigma}$ in Formula~(\ref{eq:rho}) equal to $1$ and $10$ respectively.
This model was evaluated on two datasets of personalized  models corresponding to \textit{Part-2} and  combined \textit{Part-1+Part-2} datasets.
The \textit{Indicator} dataset is the same in all experiments.

\subsubsection{Attack model A2}
\label{sssec:a2e}

For training the attack model \textbf{A2}, we use 1300 personalized  speaker models corresponding to 736 unique speakers  from \textit{Part-1}. 
When we applied the frozen part of the architecture shown in Figure~\ref{fig:a2} to the 32-minute \textit{Indicator} dataset for each speaker model in \textit{Part-1}, we obtained the training data with the amount corresponding to about 693h (32$\times$1300). 
The trained part of the NN model, illustrated in Figure~\ref{fig:a2}, has a similar topology to a conventional x-vector  extractor~\cite{snyder2018x}. However, unlike the standard NN x-vector extractor, that is trained to predict speaker id-s by the input speech segment, our proposed model  learns to predict a speaker identity from the $W_s^h$ part of a speaker personalized model.
We trained 2 attack models corresponding to the two 
values of parameter $h\in\{1,5\}$~-- a hidden layer in the ASR neural AMs at  which we compute the  activations. Values $h$ were chosen based on the results for the attack model \textbf{A1}.
The output dimension of the frozen part is 512. The frozen part is followed by the trained part that  consists of
 seven hidden TDNN layers and one statistical pooling layer  introduced after the fifth TDNN layer. The output  is a softmax layer with the targets corresponding to speakers in the pool of  speaker personalized models (number of unique speakers in \textit{Part-1}).

\subsection{Results}
\label{ssec:res}
 
The attack models were evaluated in terms of equal error rate (EER).
Denoting by $P_\text{fa}(\theta)$ and $P_\text{miss}(\theta)$ the false alarm and miss rates at threshold~$\theta$, the EER corresponds to the threshold $\theta_\text{EER}$ at which the two detection error rates are equal, i.e., $\text{EER}=P_\text{fa}(\theta_\text{EER})=P_\text{miss}(\theta_\text{EER})$.

\begin{figure}[htb]
\centering
  \centerline{\includegraphics[width=8cm]{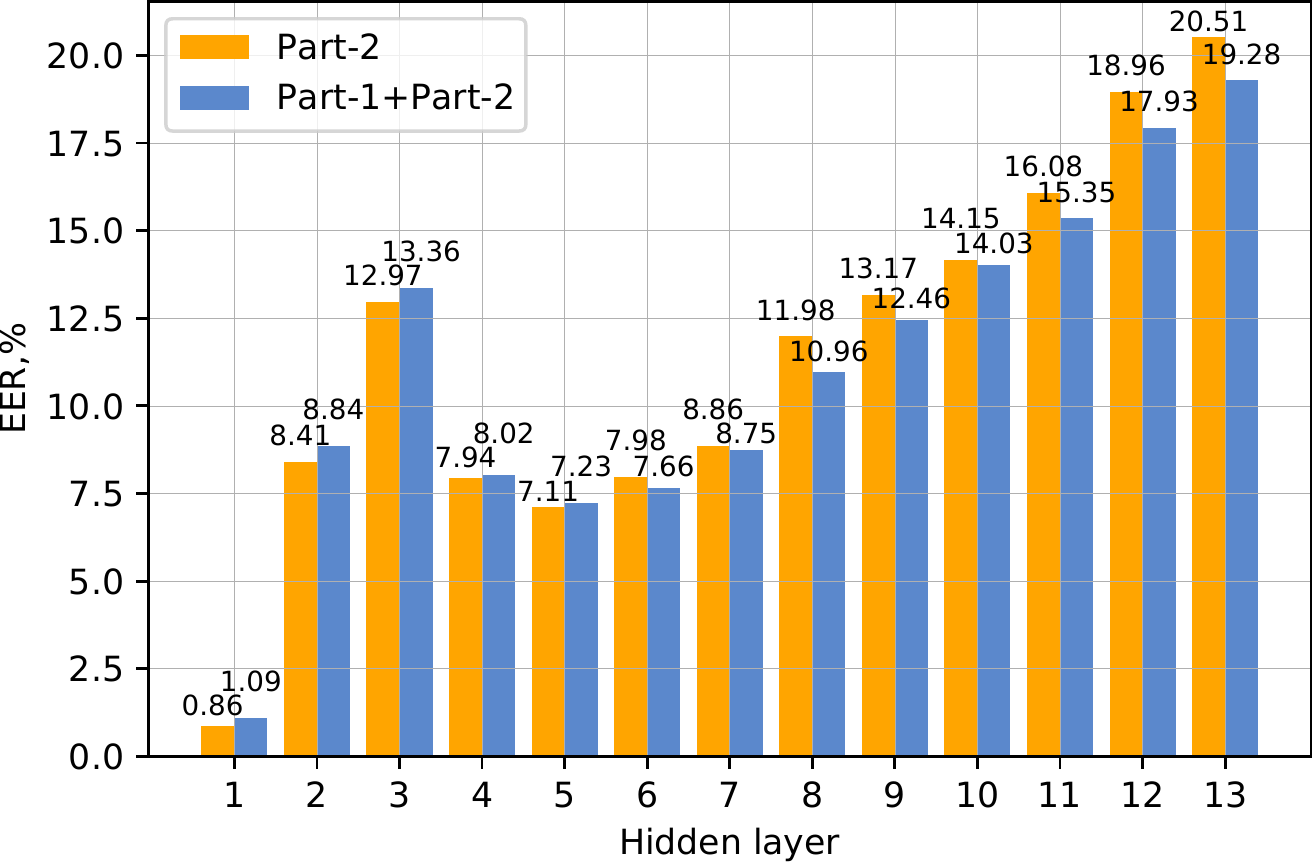}}
  \vspace{-2pt}
 \caption{EER,~\% for the attack model \textbf{A1} depending on the hidden layer $h$ (in $W_g$ and $W_{s_i}$) which was used to compute outputs,  evaluated on \textit{Part-2} and on the combined \textit{Part-1+Part-2} dataset.}
\label{fig:res}
\end{figure}

Results for the attack model \textbf{A1} are shown in Figure~\ref{fig:res} for \textit{Part-2} and combined \textit{Part-1} and \textit{Part-2} datasets.
Speaker information can be  captured for all values $h$ with varying success: EER ranges from 0.86\% (for the first hidden layer) up to 20.51\% (for the top  hidden layer) on  \textit{Part-2}. 
For the \textit{Part-1+Part-2} we observe similar results.

\vspace{-3pt}

\begin{figure}[htb]
  \centerline{\includegraphics[width=8cm]{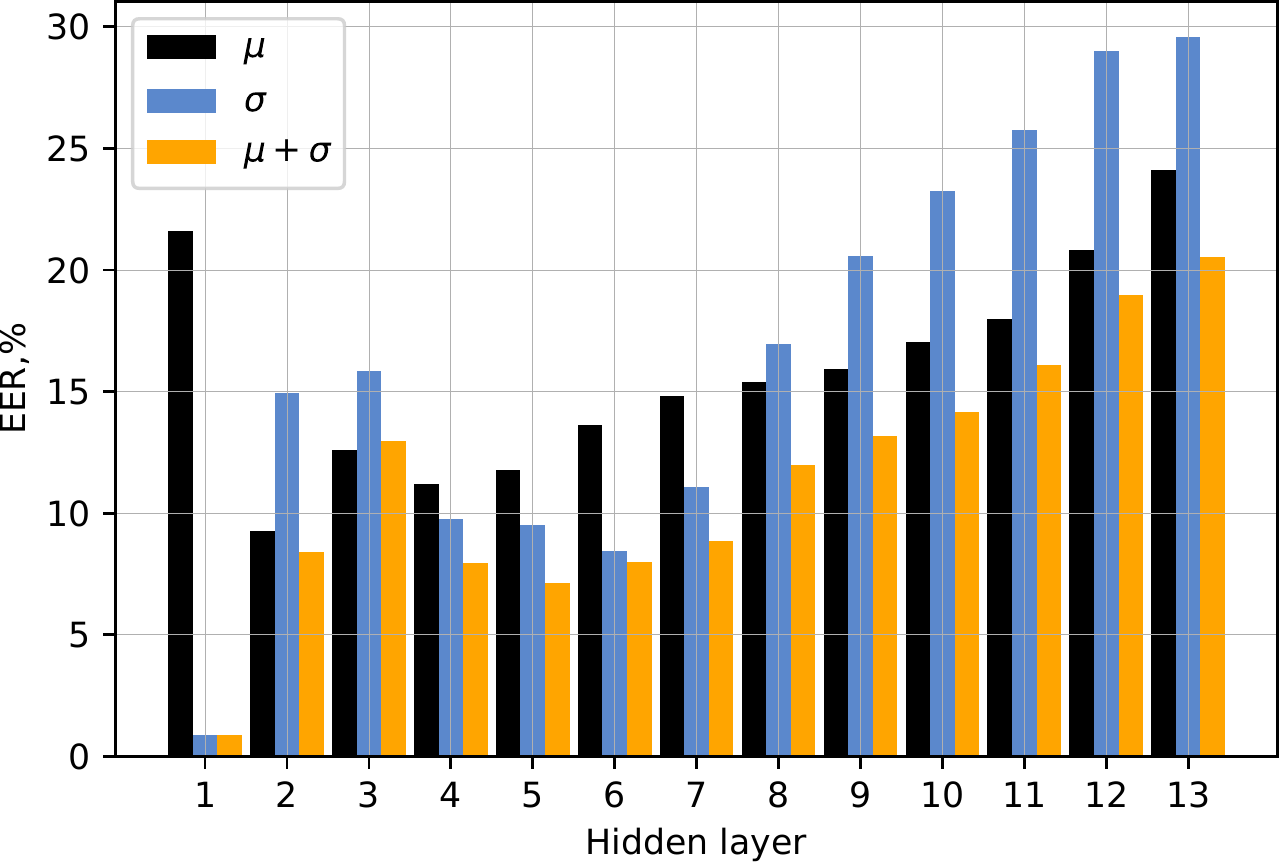}}
  \vspace{-2pt}
 \caption{EER,~\% for the attack model \textbf{A1} depending on the hidden layer $h$,  evaluated on \textit{Part-2}  dataset.  $\mu+\sigma$ -- both means and standard deviations were used to  compute similarity score $\rho$; $\mu$ -- only means; and   $\sigma$ -- only standard deviations  were used.}
\label{fig:res-a1-mu-sigma}
\end{figure}

To analyze the impact of each component in Formula~(\ref{eq:rho}) on the ASV performance, we separately compute similarity score $\rho$ either using only means ($\alpha_{\sigma}=0$) or only standard deviations ($\alpha_{\mu}=0$).
Results on the \textit{Part-2} dataset are shown in Figure~\ref{fig:res-a1-mu-sigma}. Black bars correspond to $\alpha_{\sigma}=0$ when only means were used to compute similarity scores $\rho$ between personalized models. Blue bars represent results for $\alpha_{\mu}=0$ when only standard deviations were used to compute $\rho$. Orange bars correspond to the combined usage of means and  standard deviations as in Figure~\ref{fig:res} ($\alpha_{\mu}=1$, $\alpha_{\sigma}=10$).
The impact of each component in the sum changes for different hidden layers.
When we use only standard deviations, we observe the lowest EER on the first layer. In case of using only means, the first layer is, on the contrary,  one of the  least informative for speaker verification. For all other layers,  combination of means and standard deviations provided superior results over the cases when only one of these components were used. 
These surprising results for the first hidden layer could possibly be explained by the fact that the personalized models incorporate i-vectors in their inputs, thus the speaker information can be easily learnt at this level of the NN. We plan to investigate this phenomena in detail in our future research.

We choose  two values $h\in\{1,5\}$ which demonstrate promising results for the model \textbf{A1}, and use the corresponding outputs to train two attack models with the configuration \textbf{A2}. The comparative results for the two attack models  are presented in Table~\ref{tab:a1a2}. For $h=5$, the second attack model provides significant improvement  in performance over the first one and reduces EER from 7\% down to 2\%.
For $h=1$, we could not obtain any improvement by training a NN based attack model: the results for \textbf{A2} in this case are worse than for the simple approach \textbf{A1}.
One explanation for this
phenomenon could be the following. The  first layers of the AMs provide highly informative features for speaker classification, however,  training the proposed NN model on these features results in overfitting because  training criterion of the NN is speaker accuracy, but not the target EER metric, and the number of targets is relatively small, hence, the NN overfits to classify the seen speakers in the training dataset.  

\begin{table}[ht]
  \centering
  \scalebox{1}{
  \begin{tabular}{l|r|r}
   \toprule
    \textbf{Attack model}  &  \textbf{h=1} &  \textbf{h=5}\\ \midrule
    \textbf{A1} & \textbf{0.86} & 7.11 \\
    \textbf{A2} & 12.31 & \textbf{1.94} \\
    \bottomrule
  \end{tabular}}
  \caption{EER, \% 
  evaluated on \textit{Part-2}, $h$ - indicator of a hidden layer}
  \label{tab:a1a2}
\end{table}

\vspace{-14pt}

\section{Conclusions}
\label{sec:concl}


In this work, we focused on the privacy protection problem  for ASR AMs trained in a FL framework.
We explored to what extent  ASR AMs are vulnerable to  privacy attacks.
We developed two attack models that aim to infer speaker identity from the locally updated personalized models without access to any speech data of the target speakers.
One attack model is based on the proposed similarity score between personalized AMs computed on some external \textit{Indicator} dataset, and another one is a NN   model.
We demonstrated on the TED-LIUM 3 corpus that   both  attack models are very effective and can provide EER of about 1\% for the simple attack model \textbf{A1} and  2\% for the NN  attack model \textbf{A2}.  
Another important contribution of this work is the finding that the first layer of personalized AMs contains a large amount of speaker  information
that is mainly contained in the standard deviation values computed on  \textit{Indicator} data.
This interesting property of NN adapted AMs opens new perspectives also for ASV, and in future work, we plan to use it for developing an efficient ASV system.

\bibliographystyle{IEEE}
\bibliography{mybib}

\begin{thebibliography}{10}

\bibitem{cui2021federated}
Xiaodong Cui, Songtao Lu, and Brian Kingsbury,
\newblock ``Federated acoustic modeling for automatic speech recognition,''
\newblock in {\em ICASSP}, 2021, pp. 6748--6752.

\bibitem{dimitriadis2020federated}
Dimitrios Dimitriadis, Kenichi Kumatani, Robert Gmyr, et~al.,
\newblock ``A federated approach in training acoustic models.,''
\newblock in {\em Interspeech}, 2020, pp. 981--985.

\bibitem{mdhaffar2021study}
Salima Mdhaffar, Marc Tommasi, and Yannick Est{\`e}ve,
\newblock ``Study on acoustic model personalization in a context of
  collaborative learning constrained by privacy preservation,''
\newblock in {\em Speech and Computer}, 2021, pp. 426--436.

\bibitem{guliani2021training}
Dhruv Guliani, Fran{\c{c}}oise Beaufays, and Giovanni Motta,
\newblock ``Training speech recognition models with federated learning: A
  quality/cost framework,''
\newblock in {\em ICASSP}, 2021, pp. 3080--3084.

\bibitem{dimitriadis2020dynamic}
Dimitrios Dimitriadis, Kenichi Kumatani, Robert Gmyr, et~al.,
\newblock ``Federated transfer learning with dynamic gradient aggregation,''
\newblock {\em arXiv preprint arXiv:2008.02452}, 2020.

\bibitem{yu2021federated}
Wentao Yu, Jan Freiwald, S{\"o}ren Tewes, Fabien Huennemeyer, and Dorothea
  Kolossa,
\newblock ``Federated learning in {ASR}: Not as easy as you think,''
\newblock {\em arXiv preprint arXiv:2109.15108}, 2021.

\bibitem{li2020federated}
Tian Li, Anit~Kumar Sahu, Ameet Talwalkar, and Virginia Smith,
\newblock ``Federated learning: Challenges, methods, and future directions,''
\newblock {\em IEEE Signal Processing Magazine}, vol. 37, no. 3, pp. 50--60,
  2020.

\bibitem{mothukuri2021survey}
Viraaji Mothukuri, Reza~M Parizi, Seyedamin Pouriyeh, Yan Huang, Ali
  Dehghantanha, and Gautam Srivastava,
\newblock ``A survey on security and privacy of federated learning,''
\newblock {\em Future Generation Computer Systems}, vol. 115, pp. 619--640,
  2021.

\bibitem{geiping2020inverting}
Jonas Geiping, Hartmut Bauermeister, Hannah Dr{\"o}ge, and Michael Moeller,
\newblock ``Inverting gradients--how easy is it to break privacy in federated
  learning?,''
\newblock {\em arXiv preprint arXiv:2003.14053}, 2020.

\bibitem{carlini2019secret}
Nicholas Carlini, Chang Liu, {\'U}lfar Erlingsson, Jernej Kos, and Dawn Song,
\newblock ``The secret sharer: Evaluating and testing unintended memorization
  in neural networks,''
\newblock in {\em 28th Security Symposium}, 2019, pp. 267--284.

\bibitem{truex2019demystifying}
Stacey Truex, Ling Liu, Mehmet~Emre Gursoy, Lei Yu, and Wenqi Wei,
\newblock ``Demystifying membership inference attacks in machine learning as a
  service,''
\newblock {\em IEEE Transactions on Services Computing}, 2019.

\bibitem{wang2019beyond}
Zhibo Wang, Mengkai Song, Zhifei Zhang, Yang Song, Qian Wang, and Hairong Qi,
\newblock ``Beyond inferring class representatives: User-level privacy leakage
  from federated learning,''
\newblock in {\em IEEE INFOCOM}. IEEE, 2019, pp. 2512--2520.

\bibitem{bonawitz2016practical}
Keith Bonawitz, Vladimir Ivanov, Ben Kreuter, et~al.,
\newblock ``Practical secure aggregation for federated learning on user-held
  data,''
\newblock {\em arXiv preprint arXiv:1611.04482}, 2016.

\bibitem{dwork2006differential}
Cynthia Dwork,
\newblock ``Differential privacy,''
\newblock in {\em International Colloquium on Automata, Languages, and
  Programming}, 2006.

\bibitem{pathak2013privacy}
Manas~A Pathak, Bhiksha Raj, Shantanu~D Rane, and Paris Smaragdis,
\newblock ``Privacy-preserving speech processing: cryptographic and
  string-matching frameworks show promise,''
\newblock {\em IEEE signal processing magazine}, vol. 30, no. 2, pp. 62--74,
  2013.

\bibitem{smaragdis2007framework}
Paris Smaragdis and Madhusudana Shashanka,
\newblock ``A framework for secure speech recognition,''
\newblock {\em IEEE Transactions on Audio, Speech, and Language Processing},
  vol. 15, no. 4, pp. 1404--1413, 2007.

\bibitem{xie2018differentially}
Liyang Xie, Kaixiang Lin, Shu Wang, Fei Wang, and Jiayu Zhou,
\newblock ``Differentially private generative adversarial network,''
\newblock {\em arXiv preprint arXiv:1802.06739}, 2018.

\bibitem{cohen2019voice}
Alice Cohen-Hadria, Mark Cartwright, Brian McFee, and Juan~Pablo Bello,
\newblock ``Voice anonymization in urban sound recordings,''
\newblock in {\em IEEE 29th International Workshop on Machine Learning for
  Signal Processing (MLSP)}, 2019, pp. 1--6.

\bibitem{tomashenko2020introducing}
Natalia Tomashenko, Brij Mohan~Lal Srivastava, Xin Wang, Emmanuel Vincent,
  Andreas Nautsch, Junichi Yamagishi, Nicholas Evans, et~al.,
\newblock ``Introducing the {VoicePrivacy} initiative,''
\newblock in {\em Interspeech}, 2020, pp. 1693--1697.

\bibitem{Tomashenko2021CSl}
Natalia Tomashenko, Xin Wang, Emmanuel Vincent, Jose Patino, et~al.,
\newblock ``{The VoicePrivacy 2020 Challenge: Results and findings},''
\newblock {\em Submitted to Computer Speech and Language}, 2021.

\bibitem{leroy2019federated}
David Leroy, Alice Coucke, Thibaut Lavril, Thibault Gisselbrecht, and Joseph
  Dureau,
\newblock ``Federated learning for keyword spotting,''
\newblock in {\em ICASSP}. IEEE, 2019, pp. 6341--6345.

\bibitem{hard2020training}
Andrew Hard, Kurt Partridge, Cameron Nguyen, Niranjan Subrahmanya, Aishanee
  Shah, Pai Zhu, Ignacio~Lopez Moreno, and Rajiv Mathews,
\newblock ``Training keyword spotting models on non-iid data with federated
  learning,''
\newblock {\em arXiv preprint arXiv:2005.10406}, 2020.

\bibitem{latif2020federated}
Siddique Latif, Sara Khalifa, Rajib Rana, and Raja Jurdak,
\newblock ``Federated learning for speech emotion recognition applications,''
\newblock in {\em 2020 19th ACM/IEEE International Conference on Information
  Processing in Sensor Networks (IPSN)}. IEEE, 2020, pp. 341--342.

\bibitem{granqvist2020improving}
Filip Granqvist, Matt Seigel, Rogier van Dalen, {\'A}ine Cahill, Stephen Shum,
  and Matthias Paulik,
\newblock ``Improving on-device speaker verification using federated learning
  with privacy,''
\newblock {\em arXiv preprint arXiv:2008.02651}, 2020.

\bibitem{salima2022retrieving}
Salima Mdhaffar et~al.,
\newblock ``Retrieving speaker information from personalized acoustic models
  for speech recognition,''
\newblock in {\em Submitted to ICASSP}, 2021.

\bibitem{mcmahan2017communication}
Brendan McMahan, Eider Moore, Daniel Ramage, et~al.,
\newblock ``Communication-efficient learning of deep networks from
  decentralized data,''
\newblock in {\em Artificial intelligence and statistics}. PMLR, 2017, pp.
  1273--1282.

\bibitem{snyder2018x}
David Snyder, Daniel Garcia-Romero, Gregory Sell, Daniel Povey, and Sanjeev
  Khudanpur,
\newblock ``X-vectors: Robust {DNN} embeddings for speaker recognition,''
\newblock in {\em ICASSP}. IEEE, 2018, pp. 5329--5333.

\bibitem{ioffe2006probabilistic}
Sergey Ioffe,
\newblock ``Probabilistic linear discriminant analysis,''
\newblock in {\em European Conference on Computer Vision}, 2006, pp. 531--542.

\bibitem{peddinti2015time}
Vijayaditya Peddinti, Daniel Povey, and Sanjeev Khudanpur,
\newblock ``A time delay neural network architecture for efficient modeling of
  long temporal contexts,''
\newblock in {\em Sixteenth annual conference of the international speech
  communication association}, 2015.

\bibitem{hernandez2018ted}
Fran{\c{c}}ois Hernandez, Vincent Nguyen, Sahar Ghannay, Natalia Tomashenko,
  and Yannick Est{\`e}ve,
\newblock ``{TED-LIUM~3}: twice as much data and corpus repartition for
  experiments on speaker adaptation,''
\newblock in {\em Speech and Computer}, 2018, pp. 198--208.

\bibitem{povey2011kaldi}
Daniel Povey, Arnab Ghoshal, Gilles Boulianne, Lukas Burget, et~al.,
\newblock ``The {Kaldi} speech recognition toolkit,''
\newblock in {\em ASRU}. IEEE Signal Processing Society, 2011.

\bibitem{povey2016purely}
Daniel Povey, Vijayaditya Peddinti, Daniel Galvez, Pegah Ghahremani, Vimal
  Manohar, Xingyu Na, et~al.,
\newblock ``{Purely sequence-trained neural networks for ASR based on
  lattice-free MMI},''
\newblock in {\em Interspeech}, 2016, pp. 2751--2755.

\end{thebibliography}

\end{document}